\documentclass[10pt,twocolumn,letterpaper]{article}

\usepackage[pagenumbers]{cvpr} 

\usepackage{graphicx}
\usepackage{amsmath}
\usepackage{amssymb}
\usepackage{booktabs}
\usepackage{arydshln}
\usepackage{caption}
\usepackage{subcaption}
\usepackage{bm}
\usepackage{multirow}
\usepackage{multicol}
\usepackage{adjustbox}
\usepackage{algorithm}
\usepackage{algpseudocode}

\usepackage[pagebackref,breaklinks,colorlinks]{hyperref}

\usepackage[capitalize]{cleveref}
\crefname{section}{Sec.}{Secs.}
\Crefname{section}{Section}{Sections}
\Crefname{table}{Table}{Tables}
\crefname{table}{Tab.}{Tabs.}

\def\cvprPaperID{7070} 
\def\confName{CVPR}
\def\confYear{2023}

\begin{document}

\title{Attribute-Centric Compositional Text-to-Image Generation}
\author{Yuren Cong\textsuperscript{1}, 
Martin Renqiang Min\textsuperscript{2}, 
Li Erran Li\textsuperscript{3}\thanks{The work was done outside of Amazon.},
Bodo Rosenhahn\textsuperscript{1},
Michael Ying Yang\textsuperscript{4}\\
\textsuperscript{1}\textit{TNT, Leibniz University Hannover},
\textsuperscript{2}\textit{NEC Laboratories America},\\
\textsuperscript{3}\textit{AWS AI, Amazon},
\textsuperscript{4}\textit{SUG, University of Twente}\\
}
\maketitle

\begin{abstract}
Despite the recent impressive breakthroughs in text-to-image generation, 
generative models have difficulty in capturing the data distribution of underrepresented attribute compositions while over-memorizing overrepresented attribute compositions, which raises public concerns about their robustness and fairness. To tackle this challenge, we propose \textbf{ACTIG}, an attribute-centric compositional text-to-image generation framework.
We present an attribute-centric feature augmentation  
and a novel image-free training scheme, which greatly improves model's ability to generate images with underrepresented attributes.
We further propose an attribute-centric contrastive loss to 
avoid overfitting 
to overrepresented attribute compositions.
We validate our framework on the CelebA-HQ and CUB datasets.
Extensive 
experiments show that the compositional generalization of ACTIG is outstanding, and
our framework
outperforms previous works in terms of image quality and text-image consistency.
The source code  will be released at \url{https://github.com/yrcong/ACTIG}.

\end{abstract}

\vspace{-2mm}
\section{Introduction}
\label{sec:introduction}
Recently impressive breakthroughs have been made in text-to-image generation as
several large models \cite{yu2022scaling, ramesh2022hierarchical,rombach2022high,saharia2022photorealistic} trained on large-scale datasets \cite{schuhmann2021laion,ramesh2021zero, changpinyo2021cc12m} have achieved incredible performance.
\begin{figure}[t]
  \centering
   \includegraphics[width=1\linewidth]{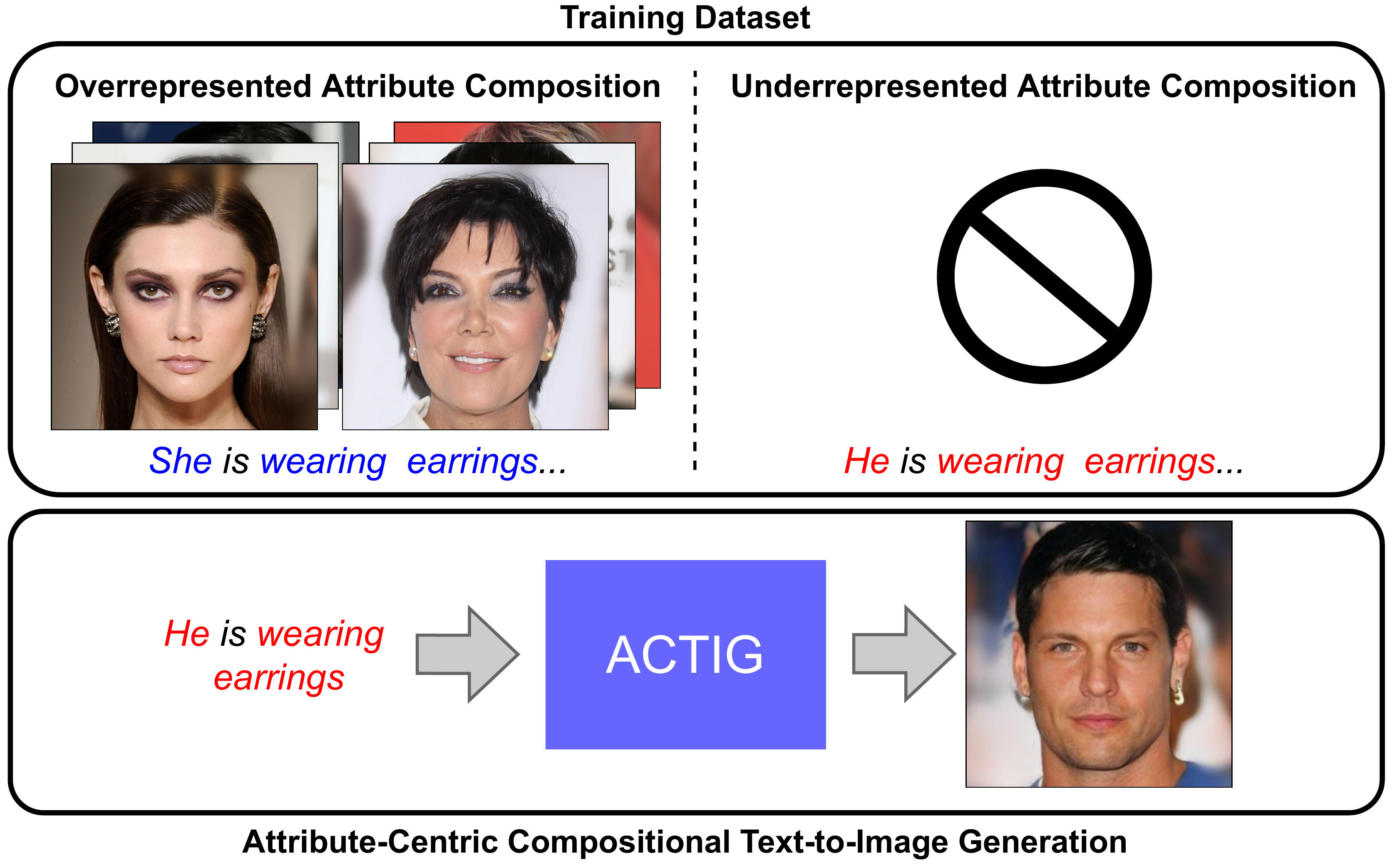}
   \caption{Our model can generate high-fidelity, text-matched images, even if the attribute compositions in the input text have not been seen in the training.}
   \label{fig:teaser}
   \vspace{-5mm}
\end{figure}
However, compositional generalization of (large) generative models
, \textit{i.e.}, the ability to compose different concepts in generation, is far from a solved problem.
It can be divided into two different 
categories: entity composition and attribute composition.
Entity composition means that generative models integrate several entities into a complex scene.
Attribute composition refers to 
combination of different attributes on individual entities.
Popular attribute compositions are not difficult for current text-to-image generation models.
However, generating an image conditional on the prompt with underrepresented attribute compositions 
remains a great challenge.

In this paper, we aim to improve
 attribute compositional generalization, for which 
there are two main challenges
: underrepresented attribute compositions and overrepresented attribute compositions.
For underrepresented attribute compositions, there are only few or no samples in the training data, while overrepresented attribute compositions have an excessive number of instances in the training data. 
It results in generative models that fail to capture the data distribution of underrepresented attribute compositions and over-memorize the features of popular compositions.
For example, in the CelebA-HQ dataset \cite{CelebAMask-HQ}, ``\textit{she}'' and ``\textit{wearing earrings}'' are a very frequent composition (see \cref{fig:teaser}), while the composition of ``\textit{he}'' and ``\textit{wearing earrings}'' is underrepresented.
The imbalanced distribution causes generative models to synthesize an image of a woman with earrings, instead of a man, when given the input ``\textit{he is wearing earrings}''.
Previous works \cite{li2022stylet2i, nie2021controllable} have attempted to solve this problem 
by manipulating attribute-directed codes in the latent space of a pre-trained generator.
However, using latent codes that do not follow the learned distribution carries the risk of 
generating images with low quality.
StyleT2I \cite{li2022stylet2i} uses spatial constraints to disentangle attributes, but an external segmentation model is necessary, which makes it difficult to 
use and extend.

A natural idea to handle underrepresented attribute compositions is to augment training samples. 
However, it is difficult to perform these image augmentations in practice. 
Fortunately, creating text with underrepresented attribute compositions is straightforward. 
Inspired by these observations, in this paper, we propose \textbf{ACTIG}, a novel attribute-centric compositional text-to-image generation framework. 
Specifically, we introduce an attribute-centric feature augmentation and a new image-free training paradigm to compensate for the data distribution.
We compute the augmented text feature using CLIP text encoder. Via our text-to-image mapping network, we obtain the augmented image features.
Our image-free training encourages the model to generate images with underrepresented attribute compositions.
%
We further propose an attribute-centric contrastive loss to disentangle the feature distribution of attributes, which 
avoids overfitting to overrepresented attribute compositions.
Our \textbf{main contributions} 
of this paper are summarized as follows:
\begin{itemize}
\item We present an attribute-centric compositional text-to-image generation framework, named ACTIG, which excels in image quality and text-image consistency.
\item We alternate between a fully supervised paradigm and a novel image-free paradigm in training, so that the model learns the feature distributions of the real data and underrepresented attribute compositions from attribute-centric feature augmentation simultaneously.
\item We propose an attribute-centric contrastive loss to disentangle the data distribution of attributes to prevent the model from over-memorizing the overrepresented attribute compositions.
\item We conduct comprehensive experiments on CelebA-HQ dataset \cite{CelebAMask-HQ} and CUB dataset \cite{wah2011caltech}, and ACTIG achieves state-of-the-art results.
\end{itemize}

\section{Related work}
\label{sec:related_work}
\noindent
\textbf{Text-to-image generation.}
Significant progress has been made in text-to-image generation over the years and a variety of models have emerged \cite{lee2022autoregressive, wu2022text, wu2022nuwa, zhou2022interactive}.
Recently, diffusion models \cite{saharia2022photorealistic, nichol2021glide, gu2022vector} trained on large-scale datasets have demonstrated tremendous promise.
DALLE2 \cite{ramesh2022hierarchical} proposes a diffusion decoder that generates an image conditioned on the CLIP embedding \cite{radford2021learning}.
Stable Diffusion \cite{rombach2022high} integrates cross-attention modules into the model structure to design powerful diffusion generators.
Auto-regressive models \cite{ramesh2021zero, gafni2022make, ding2021cogview, yu2022scaling} are also showing their potential. 
In contrast, GAN-based methods \cite{zhang2022divergan,wang2021cycle,liu2021time,crowson2022vqgan, qiao2019mirrorgan, yin2019semantics,li2019object, cheng2020rifegan, zhang2018photographic,zhu2020cookgan} have motivated many advances in text-to-image generation.
AttnGAN \cite{xu2018attngan} introduces a attentional multimodal similarity model to calculate a fine-grained  image-text correspondence loss, which is widely used in many GAN models \cite{tao2022df, liao2022text, zhu2019dm, li2019controllable, wu2022adma}.
XMC-GAN \cite{zhang2021cross} improves text-image matching through cross-modal contrastive learning.
DAE-GAN \cite{ruan2021dae} considers not only sentence-level information, but also the information of attributes extracted from the text.
As a successful image generation framework, StyleGAN \cite{karras2019style, karras2020analyzing} has also been extended for text-to-image generation.
TediGAN \cite{xia2021tedigan, xia2021towards} minimizes the embedding distances between the image and corresponding text in the latent space and employs a pre-trained StyleGAN generator to synthesize images. 
The above GANs strongly depend on the data distribution of the training set.
To improve zero-shot text-to-image generation, Lafite \cite{zhou2021lafite} proposes a language-free training framework by generating text features from image features.
However, this method is still limited by the feature distribution of entire images.
For rare attribute compositions in natural images, Lafite is unable to capture the features effectively.
Our framework, ACTIG, goes in the opposite direction.
We generate image features from attribute-centric augmented text to perform an image-free training.

\vspace{2mm}
\noindent
\textbf{Compositional image generation.}
%
A benchmark for compositional text-to-image generation \cite{park2021benchmark} is proposed, which is a study of previous text-to-image generation models for attribute compositions.
LACE \cite{nie2021controllable} proposes an energy-based model formulating attribute labels in the latent space of a pre-trained generator.
However, since the formulated latent codes do not exactly follow the learned distribution of the pre-trained generator, there is a risk of reducing the image quality.
Liu \textit{et al.} \cite{liu2022compositional} interpret diffusion models as energy-based models composing several prompts into an image. 
A spatial constraint loss is introduced in StyleT2I \cite{li2022stylet2i} to disentangle attribute features by limiting the spatial variation according to the input attributes. 
However, these works have more or less ignored the imbalanced distribution of attribute compositions in the dataset.
Our framework, ACTIG, improves the attribute compositional generalization by focusing on underrepresented and overrepresented attribute compositions.

\vspace{2mm}
\noindent
\textbf{Multi-modal representation learning.}
High-quality text representations are essential for text-to-image generation.
AttnGAN \cite{xu2018attngan} introduce a fine-grained learning framework using attention mechanism to connect words and sub-regions, while XMC-GAN obtains text embeddings from  a pre-trained BERT \cite{devlin2018bert}.
CLIP \cite{radford2021learning} is introduced to the task of text-to-image generation, and many previous works \cite{zhou2021lafite,li2022stylet2i,ramesh2022hierarchical,wang2022clip, liu2021fusedream} have demonstrated the strength of its language-and-vision feature space.
In this paper, we are inspired to pre-train a specific CLIP model connecting attributes and images to guide our attribute-centric contrastive loss, which is used to capture the independent attribute distributions in the adversarial training.

\section{Method}
\label{sec:method}
In this section, we present our framework ACTIG for attribute-centric compositional text-to-image generation.
For underrepresented attribute compositions, we propose an attribute-centric feature augmentation and an image-free training paradigm to compensate for the data distribution.
For overrepresented attribute compositions, we introduce an attribute-centric contrastive loss to capture the independent attribute distributions.

\subsection{GAN structure}
\label{sec:gan_structure}
Our generative model is built upon StyleGAN2 \cite{karras2020analyzing} with two modifications. We use a conditional generator instead.
To facilitate the image-free training, 
we adopt the original StyleGAN2  discriminator to estimate whether the 
images are real or fake.
Meanwhile, a matching discriminator using CLIP encodings is introduced to estimate the text-image consistency. 
More details about GAN structure are provided in the supplementary material.

\vspace{2mm}
\noindent
\textbf{Generator.}
The original StyleGAN2 generator consists of a mapping network and a synthesis network.
The non-linear mapping network projects the input latent code $\bm{z}$ into a latent space $\mathcal{W}$, while the synthesis network generates images based on the output of the mapping network.
To make the unconditional generator conditional, we normalize and concatenate the latent code $\bm{z}$ and text encoding $\bm{t}$ provided by the CLIP encoder as input to the mapping network, while the generator architecture remains unchanged.
Therefore, an image $\hat{I}$ generated by the generator $G$ can be formulated as: $\hat{I}=G(\bm{z}, \bm{t})$.

\vspace{2mm}
\noindent
\textbf{Discriminators.}
Different from previous works \cite{ruan2021dae,zhou2021lafite,liao2022text,tao2022df} that use a shared discriminator backbone to perform the tasks of estimating 
photo-fidelity
and text-image consistency simultaneously, we adopt a fidelity discriminator $D_f$ and a matching discriminator $D_m$, which are independent of each other.
This allows us to update them in a flexible way for image-free training.
For the fidelity discriminator $D_f$, we directly use the StyleGAN2 discriminator.
For the matching discriminator $D_m$, we utilize the CLIP text encoder $E_{\textrm{txt}}$ and image encoder $E_{\textrm{img}}$ to compute the encodings of the input text and image.
Two 
fully-connected layers transform the text and image encodings respectively, while the cosine similarity between the transformed embeddings is calculated to represent the text-image consistency.

\subsection{Text-to-image mapping}
\label{sec:text_to_image_mapping}
We propose a text-to-image mapping network which 
projects a CLIP text encoding $\bm{t}$ into CLIP image space to obtain the approximate CLIP image encoding $\Tilde{\bm{i}}$, which 
is later used for image-free training.
The mapping network consists of multiple 
fully-connected layers with residual connection and batch normalization.
We show the architecture in the supplementary material.
The input and output dimensions are the same as the CLIP encoding dimension.
The mapping network is pre-trained before optimizing GAN.
We combine three loss functions to enforce $\Tilde{\bm{i}}$ 
close to the real image encoding $\bm{i}$ from different views.
Mean squared error and cosine similarity loss are adopted to align $\Tilde{\bm{i}}$ and $\bm{i}$ in Euclidean space and cosine space, respectively.
We also use a contrastive loss to make $\Tilde{\bm{i}}$ most similar to the corresponding $\bm{i}$ in the batch.
Given a batch of $N$ text-image pairs, the complete objective function for the mapping network can be presented as,
\begin{equation}
\centering
\begin{aligned}
&L_{mapping} = L_{MSE}+L_{similarity}+L_{contrast}, \\
&L_{MSE}=\frac{1}{N} \sum_{k=1}^N (\Tilde{\bm{i}}_k-\bm{i}_k)^2,\\
&L_{similarity}=\frac{1}{N} \sum_{k=1}^N (1-\frac{\Tilde{\bm{i}}_k\cdot\bm{i}_k}{\left \|\Tilde{\bm{i}}_k\right \|\left \|\bm{i}_k\right \|}),\\
&L_{contrast}=-\frac{1}{N} \sum_{k=1}^N \log\frac{\exp(\textrm{sim}(\Tilde{\bm{i}}_k, \bm{i}_k)}{\sum_{j=1}^N \exp(\textrm{sim}(\Tilde{\bm{i}}_k, \bm{i}_j)},
\end{aligned}
\label{eq:text_image_mapping}
\end{equation}
where $\bm{i}$ indicates the image encoding inferred from the real image and $\Tilde{\bm{i}}$ is the mapping network output.
$\textrm{sim}(.,.)$ denotes the cosine similarity in our paper.

\begin{figure}[t]
\centering
\includegraphics[width=1\linewidth]{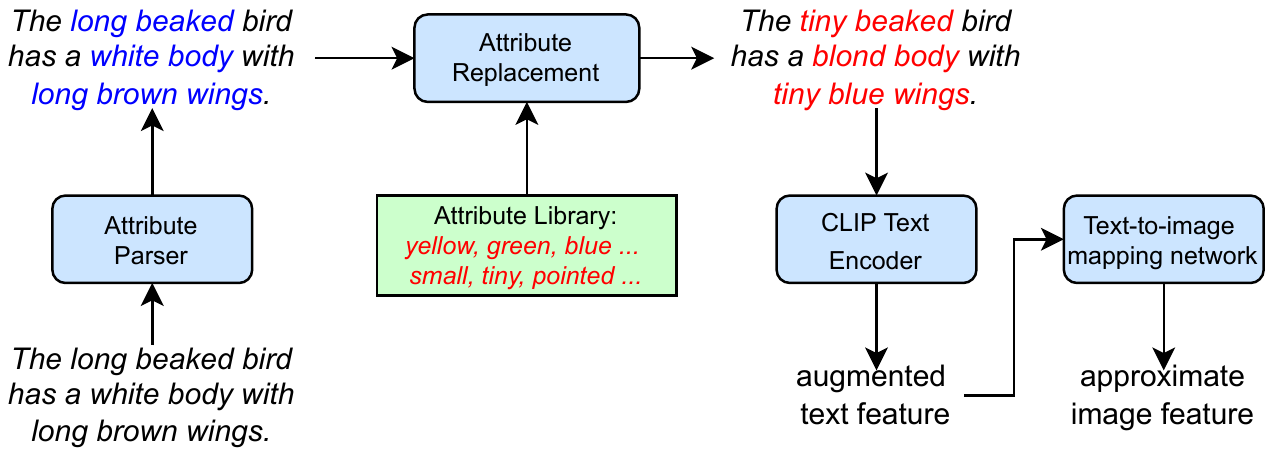}
\caption{Overview of our attribute-centric feature augmentation (for CUB \cite{wah2011caltech}). We construct an attribute library and replace the attributes in the text with those randomly sampled from the library.}
\label{fig:attribute_centric_feature_augmentation}
\vspace{-3mm}
\end{figure}

\begin{figure*}[ht!]
\centering
\includegraphics[width=0.95\linewidth]{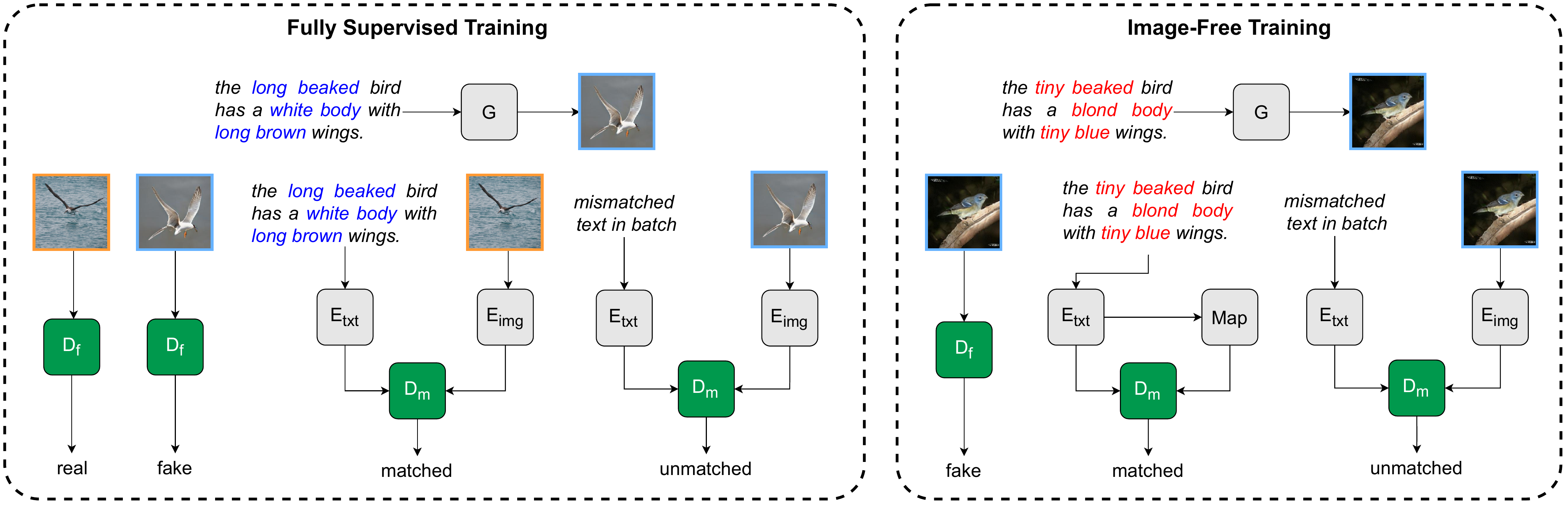}
\caption{Update paradigm for discriminators in (1) fully supervised training and (2) image-free training. The prompt with blue attributes is ground truth text in the training set, while the prompt with red attributes is produced by replacing the attributes in the left prompt with other random attributes. Map indicates our pre-trained text-to-image mapping network. We highlight the activated modules in green.}
   \label{fig:discriminator_training}
   \vspace{-3mm}
\end{figure*}

\subsection{Attribute-centric feature augmentation}
\label{sec:attribute_centric_feature_augmentation}
For better quality and text correspondence of images generated based on the underrepresented attribute compositions, an intuitive idea is adding more training samples for these compositions.
Although image augmentation is almost impossible, it is feasible to augment the text.
We generate the text with underrepresented attribute compositions by randomly selecting attributes to form prompts (for CelebA-HQ \cite{CelebAMask-HQ}) or by replacing attributes in the training prompts with other randomly sampled attributes (for CUB \cite{wah2011caltech}).
The generated prompts are encoded by CLIP text encoder and our text-to-image mapping network transforms the text encodings to the approximate image encodings.
The augmented feature pairs are utilized in the image-free training.
The attribute-centric feature augmentation pipeline for CUB is shown in Fig. \ref{fig:attribute_centric_feature_augmentation}.
The augmentation pipeline for CelebA-HQ and more details of the attribute parser such as attribute library are provided in the supplementary material.

\subsection{Attribute-centric contrastive loss}
\label{sec:attribute_centric_contrastive_loss}
To prevent overfitting 
to overrepresented attribute compositions and disentangle their distributions, we propose an attribute-centric contrastive loss that leverages the generality and transferability of CLIP.
We first finetune the pre-trained CLIP in a conventional manner, except that the training data are not text-image pairs.
In each iteration of training, we randomly sample an attribute from the text, and form an attribute-image pair with the corresponding image to replace the text-image pair.
To distinguish 
the CLIP,
which is fine-tuned with attribute-image pairs, from the text-image CLIP, we call it CLIP-A.
Given a batch of $N$ text-image pairs, we randomly extract an attribute $A$ from each text $T$.
The attribute-centric contrastive loss for $k$-th attribute-image pair can be formulated as,
\begin{equation}
\vspace{-1mm}
\centering
\begin{aligned}
L_{attr}=- \log\frac{\exp( \textrm{sim}(E_{\textrm{CLIP-A}}^{\textrm{img}}(I_k),E_{\textrm{CLIP-A}}^{\textrm{txt}}(A_k))}{\sum_{j=1}^N \exp( \textrm{sim}(E_{\textrm{CLIP-A}}^{\textrm{img}}(I_k),E_{\textrm{CLIP-A}}^{\textrm{txt}}(A_j)))} \\
- \log\frac{\exp( \textrm{sim}(E_{\textrm{CLIP-A}}^{\textrm{img}}(I_k),E_{\textrm{CLIP-A}}^{\textrm{txt}}(A_k))}{\sum_{j=1}^N \exp( \textrm{sim}(E_{\textrm{CLIP-A}}^{\textrm{img}}(I_j),E_{\textrm{CLIP-A}}^{\textrm{txt}}(A_k)))}, 
\end{aligned}
\label{eq:clip_pretrain_loss}
\end{equation}
where $I$ indicates the image. $E_{\textrm{CLIP-A}}^{\textrm{img}}$ and $E_{\textrm{CLIP-A}}^{\textrm{txt}}$ are the image encoder and text encoder of CLIP-A. 
This loss function is used in the adversarial training to capture the independent attribute distributions and  avoid the generative model over-memorizing the popular attribute compositions.

\begin{figure*}[t]
  \centering
   \includegraphics[width=0.99\linewidth]{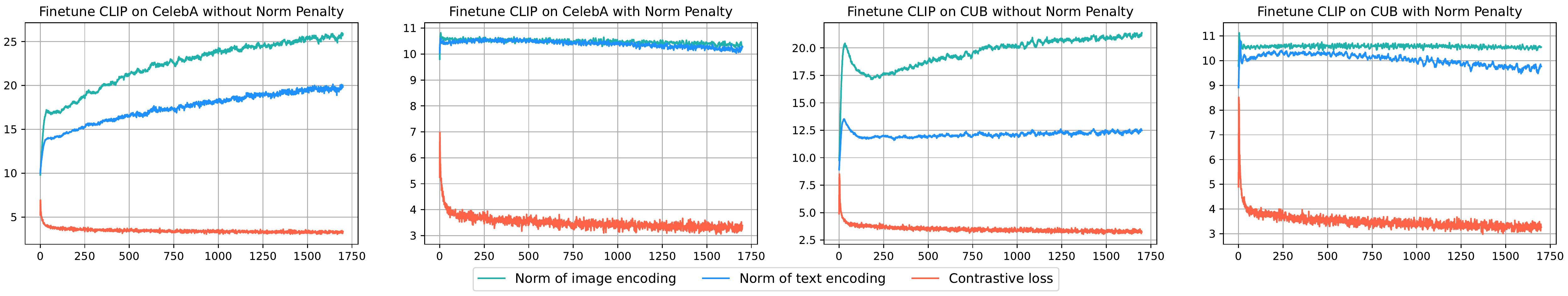}
   \caption{$\ell^2$ norms of text encoding and image encoding from the pre-trained CLIP are essentially identical. They increase at different rates during the finetuning. We introduce a norm penalty to keep the norms at the same level, which facilitates our text-to-image mapping.}
   \label{fig:pretraining_clip}
   \vspace{-3mm}
\end{figure*}

\subsection{Training schemes}
\label{sec:training_script}
Since it is 
extremely difficult to directly augment the text-image pairs of underrepresented attribute compositions required for the standard adversarial training, we train our generative model in two paradigms alternatively: (1) \textbf{fully supervised training}, which uses text-image pairs in the training set, and (2) \textbf{image-free training}, which uses the feature pairs resulting from the attribute-centric feature augmentation.
The model learns the distributions of the real data and underrepresented attribute compositions.
The training schedule of ACTIG is summarized in \cref{alg:ACTIG}.

For both training paradigms, the generator is updated in the same way. Given a batch of $N$ text, the standard unconditional loss for the generator can be presented as,
\begin{equation}
\centering
\begin{aligned}
L_{G}^f= \frac{1}{N} \sum_k^N \zeta(-D_f(\hat{I_k})), 
\end{aligned}
\label{eq:g_dr_loss}
\vspace{-2mm}
\end{equation}
where $\zeta$ denotes the softplus function and $\hat{I}$ denotes the generator output.
To 
match the generated images to the input text, we introduce a matching loss using the matching discriminator $D_m$,
\begin{equation}
\vspace{-2mm}
\centering
\begin{aligned}
L_{G}^m= \frac{1}{N} \sum_k^N(1 - D_m(\bm{t}_k, \hat{\bm{i}}_k)), 
\end{aligned}
\label{eq:g_dm_loss}
\end{equation}
where $\bm{t}$ and $\hat{\bm{i}}$ are the text and generated image embeddings respectively provided by the CLIP text encoder $E_{\textrm{txt}}$ and image encoder $E_{\textrm{img}}$.
Furthermore, we adopt the CLIP-guided contrastive loss from \cite{li2022stylet2i,zhou2021lafite},
\begin{equation}
\centering
\begin{aligned}
L_{G}^{const}=-\frac{1}{N} \sum_{k=1}^N (\log\frac{\exp( \textrm{sim}(\hat{\bm{i}}_k, \bm{t}_k))}{\sum_{j=1}^N \exp( \textrm{sim}(\hat{\bm{i}}_k, \bm{t}_j))}\\
+ \log\frac{\exp( \textrm{sim}(\hat{\bm{i}}_k, \bm{t}_k))}{\sum_{j=1}^N \exp( \textrm{sim}(\hat{\bm{i}}_j, \bm{t}_k))}).
\end{aligned}
\label{eq:g_image_contrast_loss}
\end{equation}
In order to align the generated image features with the attribute features in CLIP-A feature space, we integrate the attribute-centric contrastive loss $L_{attr}$.
The complete objective function for updating the generator is,
\begin{equation}
\centering
\begin{aligned}
L_{G}= L_G^f+L_G^m+L_G^{const}+L_{attr}.
\end{aligned}
\label{eq:sum_g_loss}
\end{equation}

In fully supervised training and image-free training, the discriminators are updated in different ways (see Fig.~\ref{fig:discriminator_training}).
For fully supervised training, given a batch of $N$ text-image pairs, the loss function for updating the fidelity discriminator is calculated as,
\begin{equation}
\centering
\begin{aligned}
L_{D}^f= \frac{1}{N} \sum_{k=1}^N (\zeta(-D_f(I_k)) + \zeta(D_f(\hat{I_k}))),
\end{aligned}
\label{eq:fully_dr_loss}
\end{equation}
while for the matching discriminator,
\begin{equation}
\centering
\begin{aligned}
L_{D}^m= \frac{1}{N} \sum_{k=1}^N (1 - D_m(\bm{t}_k, \bm{i}_k) +  D_m(\bm{t}_{k*}, \hat{\bm{i}}_{k})),
\end{aligned}
\label{eq:full_dm_loss}
\end{equation}
where $\bm{t}_{k}$ is the CLIP text encoding of the $k$-th prompt, while $\bm{t}_{k*}$ is the CLIP text encoding of a mis-matched prompt in the batch.
Therefore, the complete objective function for updating the discriminators in fully supervised training is computed as,
\begin{equation}
\centering
\begin{aligned}
L_{D}= L_{D}^f+L_{D}^m.
\end{aligned}
\label{eq:fully_D_loss}
\end{equation}
For image-free training, only attribute-centric augmented text are available.
Therefore we skip the first term in \cref{eq:fully_dr_loss} and only update the fidelity discriminator based on the generated images,
\begin{equation}
\centering
\begin{aligned}
\Tilde{L_{D}^f}=  \frac{1}{N} \sum_{k=1}^N \zeta(D_f(\hat{I_k})).
\end{aligned}
\label{eq:loss_dr_imagefree}
\end{equation}
For updating the matching discriminator, we use the pre-trained text-to-image mapping network to transform the text encodings $\bm{t}$ of augmented text to approximate image encodings $\Tilde{\bm{i}}$.
The matching discriminator loss is formulated as,
\begin{equation}
\centering
\begin{aligned}
\Tilde{L_{D}^m}=  \frac{1}{N} \sum_{k=1}^N (1 - D_m(\bm{t}_k, \Tilde{\bm{i}_k}) +  D_m(\bm{t}_{k*}, \hat{\bm{i}}_{k})).
\end{aligned}
\label{eq:loss_dm_imagefree}
\end{equation}
The complete objective function for updating the discriminators in image-free training is presented as,
\begin{equation}
\centering
\begin{aligned}
\Tilde{L_{D}}= \Tilde{L_{D}^f}+\Tilde{L_{D}^m}.
\end{aligned}
\label{eq:loss_d_imagefree}
\end{equation}
\vspace{-2mm}
\begin{algorithm}
\caption{Training schedule of ACTIG}
\label{alg:ACTIG}
\begin{algorithmic}
\State \textbf{Input:} Dataset with text-image pairs $\{\bm{T},\bm{I}\}$
\While{not converge}
\State sample mini-batch $\{T_i,I_i\}_{i=1}^N$;
\State sample latent code $\{\bm{z}_i\}_{i=1}^N$
\State// \textit{\textcolor{red}{Image-free activated in 1 out of every 4 iterations.}}
\If{fully supervised training} 
\State Generate images $\{\hat{I}_i\}_{i=1}^N$ $=G(\bm{z}, \bm{t})$;
\State Update $G$ with \cref{eq:sum_g_loss};
\State Update $D_f$ and $D_m$ with \cref{eq:fully_D_loss}    
\ElsIf{image-free training}
\State Implement attribute-centric feature augmentation
\State// \textit{\textcolor{red}{$\bm{t}$ is updated with the augmented text.}}
\State Generate images $\{\hat{I}_i\}_{i=1}^N$ $=G(\bm{z}, \bm{t})$;
\State Update $G$ with \cref{eq:sum_g_loss};
\State Update $D_f$ and $D_m$ with \cref{eq:loss_d_imagefree}
\EndIf
\EndWhile
\end{algorithmic}
\end{algorithm}

\subsection{Finetuned CLIP}
\label{sec:finetuned_clip}
We finetune the pre-trained CLIP \cite{radford2021learning}, respectively with text-image pairs and attribute-image pairs in the training set of CelebA-HQ and CUB.
CLIP trained with text-image pairs provides the text and image encodings for the generator, matching discriminator, text-image mapping network and CLIP-guided contrastive loss.
The model finetuned with attribute-image pairs, CLIP-A, is adopted for our attribute-centric contrastive loss.
However, we observe that the $\ell^2$ norms of text encoding and image encoding increase at different rates during the finetuning stage as shown in Fig. \ref{fig:pretraining_clip}.
This obviously complicates the projection of text encodings into the image encoding space and makes it difficult for the text-to-image mapping network to infer the approximate image encodings.
Therefore, we add a norm penalty to the objective function to keep the norms of text encoding and image encoding at the same value level,
\begin{equation}
\centering
\begin{aligned}
L_{\textrm{norm}}=\sigma(\|E_{\textrm{img}}(I)\|_2-\tau) + \sigma(\|E_{\textrm{txt}}(T)\|_2-\tau),
\end{aligned}
\label{eq:clip_pretrain_loss}
\end{equation}
where $\sigma$ denotes ReLU operation and $\tau$ is a threshold hyperparameter for norm penalty.


\section{Experiments}
\label{sec:experiments}
\subsection{Datasets}
\label{sec:datasets}
We train and validate our model on two datasets, CelebA-HQ \cite{CelebAMask-HQ} and CUB \cite{wah2011caltech}.
CelebA-HQ is a large-scale face dataset with facial attributes. We use the data split and text annotations proposed by Xia \textit{et al.} \cite{xia2021tedigan} and Li \textit{et al.} \cite{li2022stylet2i}, while there are $23.4k$ images for training and $1.9k$ images for testing. 
To evaluate the compositional ability, only captions with unseen attribute compositions are retained in the test set.
CUB is a dataset that includes $11.8k$ images in 200 bird species.
For text annotations, we use the captions provided by Reed \textit{et al.} \cite{reed2016learning}. 
The settings from StyleT2I \cite{li2022stylet2i} are adopted for a fair comparison.

\subsection{Evaluation metrics}
\label{sec:evaluation_metrics}

\noindent
\textbf{FID.}
We adopt Fr\'{e}chet Inception Distance (FID) \cite{heusel2017gans} to evaluate the quality of the generated images.
FID computes the distance between the feature distributions of the generated images and real images.
Lower number denotes that the synthetic images are more realistic.

\vspace{2mm}
\noindent
\textbf{R-Precision.}
We adopt R-precision \cite{xu2018attngan} to evaluate the consistency of the input text and output images.
R-Precision calculates the top 1 retrieval accuracy when using the generated image as a query to retrieve matching text from $K$ candidate texts.
If not otherwise specified, the default value of $K$ is 100.
We follow \cite{park2021benchmark} to calculate R-Precision using the CLIP finetuned on the whole dataset, which has been demonstrated to be closer to human perception.

\vspace{2mm}
\noindent
\textbf{User study.}
Although the above model-based evaluation metrics can indicate the quality of the generated images and text-image consistency, they cannot be fully equated with human perception.
Therefore, we invited $12$ participants to perform the user study on both datasets to estimate image quality and text-image consistency as Li \textit{et al.} \cite{li2022stylet2i} did. 
Given an input text, participants were requested to rank the images synthesized by different models 
on image quality and image-text consistency respectively. 

\subsection{Implementation details}
\noindent
\textbf{GAN details.}
We adopt the generator and discriminator of StyleGAN2 \cite{karras2020analyzing}. 
The input dimension of the generator is 1024, since we concatenate the CLIP text encodings to latent codes as input. 
The resolution of generated images is set to $256\times256$.
We integrate the pre-trained CLIP encoders into the matching discriminator which are frozen while training the GAN.
For both datasets, we train the generator and discriminators from scratch on 8 GPUs with Adam \cite{kingma2014adam} setting the batch size to 8 per GPU.
We alternate between fully supervised training and image-free training.
Three out of every four iterations use fully supervised training and one image-free training. The text-to-image mapping network, CLIP and CLIP-A are frozen at this stage.

\vspace{2mm}
\noindent
\textbf{Text-to-image mapping network details.}
The text-to-image mapping network consist of three linear transformation
layers with GeLU activation. 
There are residual connections between the transformation layers.
The input text encodings and target image encodings are provided by the finetuned CLIP.
We train the text-to-image mapping network with Adam \cite{kingma2014adam} setting the batch size 128 (for CelebA-HQ) and 256 (for CUB).

\vspace{2mm}
\noindent
\textbf{Finetuned CLIP details.}
To ensure the proper working of ACTIG, we finetune the pre-trained CLIP (ViT-B/32) with text-image pairs and attribute-image pairs in the training split of CelebA-HQ and CUB.
The last few layers of CLIP are finetuned according to \cite{li2022stylet2i}.
The hyperparameter $\tau$ in \cref{eq:clip_pretrain_loss} is set to $10$.
Furthermore, we finetune the pre-trained CLIP with text-image pairs from the full dataset of CelebA-HQ and CUB, denoted as CLIP-Eval.
Note that CLIP-Eval is not involved in any training and is only available for calculating R-Precision.

\begin{figure*}[ht]
  \centering
   \includegraphics[width=0.95\linewidth]{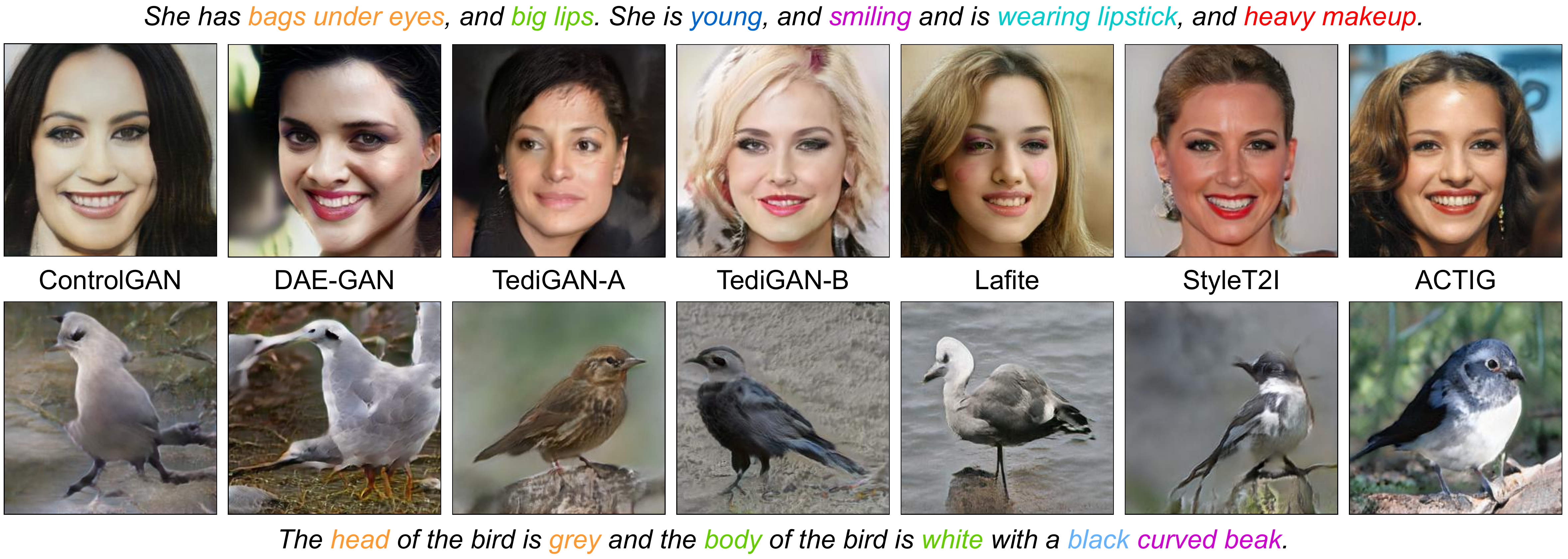}
   \caption{Qualitative comparison of different text-to-image generation models on CelebA-HQ and CUB datasets.}
   \label{fig:qualitative}
   \vspace{-3mm}
\end{figure*}

\subsection{Quantitative results and comparison}

\begin{table}[t!]
\centering
\begin{adjustbox}{width=0.49\textwidth}
\begin{tabular}{c|cc|cc}
 \hline \hline
 \multirow{2}*{Method} & \multicolumn{2}{c|}{CelebA-HQ} & \multicolumn{2}{c}{CUB} \\
  & FID $\downarrow$ & R-Precision $\uparrow$ & FID $\downarrow$ & R-Precision $\uparrow$ \\
  \hline 
ControlGAN \cite{li2019controllable} &  31.4 & 43.5 & 29.0& 13.7\\
DAE-GAN \cite{ruan2021dae} &  30.7 & 48.4 & 27.0 & 14.5\\
TediGAN-A \cite{xia2021tedigan}& 16.5 & 4.4 & 16.4 & 7.1\\
TediGAN-B \cite{xia2021towards} &  \textbf{15.5} & 30.6 & 16.8& 12.1\\
Lafite \cite{zhou2021lafite} &  17.2 & 56.9 & 15.1 & 20.7\\
StyleT2I \cite{li2022stylet2i} &  17.5 & 62.5 & 20.5& 26.4\\
ACTIG (Ours) & 15.6 & \textbf{65.0} & \textbf{12.5} & \textbf{28.7}\\
\hline \hline
\end{tabular}
\end{adjustbox}
\vspace{-2mm}
\caption{Results of text-to-image generation on CelebA-HQ and CUB datasets. A lower FID indicates better image quality, while a higher R-Precision indicates better text-image consistency.}
\label{tab:quantitative_result}
\end{table}

\begin{table}[t!]
\centering
\begin{adjustbox}{width=0.5\textwidth}
\begin{tabular}{c|ccc|ccc}
 \hline \hline
 Attribute  & \multicolumn{3}{|c}{CelebA-HQ}  & \multicolumn{3}{|c}{CUB}\\
 number &$K=10$ & $K=50$ & $K=100$  &$K=10$ & $K=50$ & $K=100$ \\
  \hline 
$\leqslant 2$ & 81.6 & 52.6 & 41.3  & 71.3 & 39.1 & 26.0\\
3 & 93.4 & 78.3 & 67.0 & 71.5 & 39.3 & 26.0\\
4 & 91.1 & 73.9 & 63.3 & 72.7 & 41.1 & 30.1\\
5 & 92.7 & 77.1 & 67.5 & 72.9 & 43.0 & 31.4\\
$ \geqslant 6$& 91.5 & 74.6 & 63.8 & 71.8 & 42.8 & 31.2\\
\hline \hline
\end{tabular}
\end{adjustbox}
\vspace{-2mm}
\caption{R-Precision of ACTIG on the CelebA-HQ and CUB datasets for different number of attributes in the input text. $K$ denotes the number of candidate text in retrieval.} 
\label{tab:attribute_num}
\vspace{-3mm}
\end{table}

\cref{tab:quantitative_result} shows the comparison between advanced text-to-image generation methods on the CelebA-HQ and CUB datasets.
For both CelebA-HQ and CUB datasets, our framework, ACTIG, achieves state-of-the-art performance.
In terms of R-Precision, ACTIG is 2.5 and 2.3 higher than another compositional text-to-image generation model StyleT2I \cite{li2022stylet2i} on CelebA-HQ and CUB, respectively.
In addition, our end-to-end trained ACTIG has a significant advantage in the image quality compared to StyleT2I, which performs on a pre-trained StyleGAN2.
We also reproduce a language-free model Lafite \cite{zhou2021lafite} on two datasets.
On both datasets, the end-to-end trained Lafite has higher FID scores than StyleT2I, however its generated images do not match the input text as well as StyleT2I without considering composition generalization.
Our end-to-end ACTIG focuses on text-to-image consistency while keeping the output image to be high-fidelity.
The FID score of ACTIG on the CelebA-HQ dataset is very close to that of TediGAN-B. 
Most of the other models also have FID between 15 and 18.
This contradicts the results of human evaluation of image quality in user study.
We conjecture that such contradiction  may be caused by the small number of test images in the test  split, whose feature distribution differs from the training split.

To demonstrate the attribute compositional generalization of ACTIG, we evaluate the text-image consistency given input prompts with different number of attributes (see \cref{tab:attribute_num}).
R-Precision is low when the attribute number in the input text is small.
This is a result of the brief descriptions leading to the lack of obvious semantic features in the generated images.
R-Precision increases as the number of constraints in the text increases.
It shows that ACTIG can generate images that match the text including complex attribute compositions.
When the attribute number is more than 6, there is a slight drop in R-Precision, possibly due to too much complex semantic information.

\subsection{Ablation study}
\label{sec:ablation_study}
\begin{table}[t!]
\centering
\begin{adjustbox}{width=0.49\textwidth}
\begin{tabular}{ccc|cc}
\hline \hline
$D_m$ & $L_{attr}$ & Image-free training & FID $\downarrow$  &R-Precision $\uparrow$ \\
\hline
- &- &- & 18.3 & 15.2 \\
\checkmark&- &- & 13.9  & 21.8\\
\checkmark &\checkmark &- & 12.8  & 25.6\\
\checkmark &\checkmark &\checkmark &\textbf{12.5} & \textbf{28.7} \\
\hline \hline
\end{tabular}
\end{adjustbox}
\vspace{-2mm}
\caption{Results for the ablation study of ACTIG on the CUB dataset. $\checkmark$ indicates the corresponding component is activated.}
\label{tab:ablate_modules}
\vspace{-2mm}
\end{table}

\begin{table}[t!]
\centering
\begin{adjustbox}{width=0.49\textwidth}
\begin{tabular}{c|cc|cc}
 \hline \hline
 \multirow{2}*{CLIP } & \multicolumn{2}{c|}{CelebA-HQ} & \multicolumn{2}{c}{CUB} \\
  & FID $\downarrow$ & R-Precision $\uparrow$ & FID $\downarrow$ & R-Precision $\uparrow$ \\
  \hline 
w/o Finetune &  16.2 & 57.1 & 13.1 & 19.5\\
Finetune w/o Norm Penalty &  15.8 & 61.2 & \textbf{12.3} & 28.2\\
Finetune w/ Norm Penalty & \textbf{15.6} & \textbf{65.0} & 12.5 & \textbf{28.7}\\
\hline \hline
\end{tabular}
\end{adjustbox}
\vspace{-2mm}
\caption{Performance of ACTIG using different CLIP models.}
\label{tab:ablate_clip}
\vspace{-3mm}
\end{table}

\noindent
\textbf{Matching discriminator.}
Many previous works \cite{ruan2021dae,zhou2021lafite,liao2022text,tao2022df} use a shared discriminator backbone to estimate fidelity and text-image consistency simultaneously.
The discriminator backbone provides the image feature to two sub-networks, one of which converts the features into a scalar representing the truthfulness, and the other combines the feature and text encoding to output the degree of text-image consistency.
In contrast, we adopt an independent CLIP-based matching discriminator $D_m$. 
We compare the model performance using the shared discriminator backbone (first row in \cref{tab:ablate_modules}) and independent $D_m$ (second row in \cref{tab:ablate_modules}) on the CUB dataset.
The results show that both image quality and text-image consistency are significantly improved with the independent $D_m$.

\vspace{2mm}
\noindent
\textbf{Attribute-centric contrastive loss.}
We further validate the effect of the attribute-centric contrastive loss $L_{attr}$ (third row in \cref{tab:ablate_modules}).
With the integration of the loss function, the quality of generated images improves and the FID decreases by 1.1.
Meanwhile, R-precision increases from 21.8 to 25.6.
It demonstrates that the attribute-centric contrastive loss helps the generator to capture the independent feature distribution for each attribute during training, while having less impact on the feature distribution of the  images.

\vspace{2mm}
\noindent
\textbf{Image-free training.}
We activate the image-free training which is supported by the attribute-centric feature augmentation. 
The fourth row in \cref{tab:ablate_modules} denotes the performance of full ACTIG.
The compositional generalization of ACTIG is further improved by introducing new attribute compositions and corresponding approximated image encodings into the image-free training.
The FID score does not have a significant change since no real image is actually entered.

\vspace{2mm}
\noindent
\textbf{CLIP finetuning.}
To verify the impact of CLIP on text-to-image mapping in \cref{sec:text_to_image_mapping} and the overall framework, we test three different CLIPs, namely the pre-trained CLIP (ViT-B/32), the CLIP finetuned without norm penalty and the CLIP finetuned with norm penalty.
The results are shown in \cref{tab:ablate_clip}.
For both datasets, while ACTIG using the pre-trained CLIP has the lowest performance, 
ACTIG using the CLIP finetuned with norm penalty has the best R-Precision, since the text-to-image mapping is facilitated.

\subsection{Qualitative results}
\begin{figure}[http]
  \centering
   \includegraphics[width=0.95\linewidth]{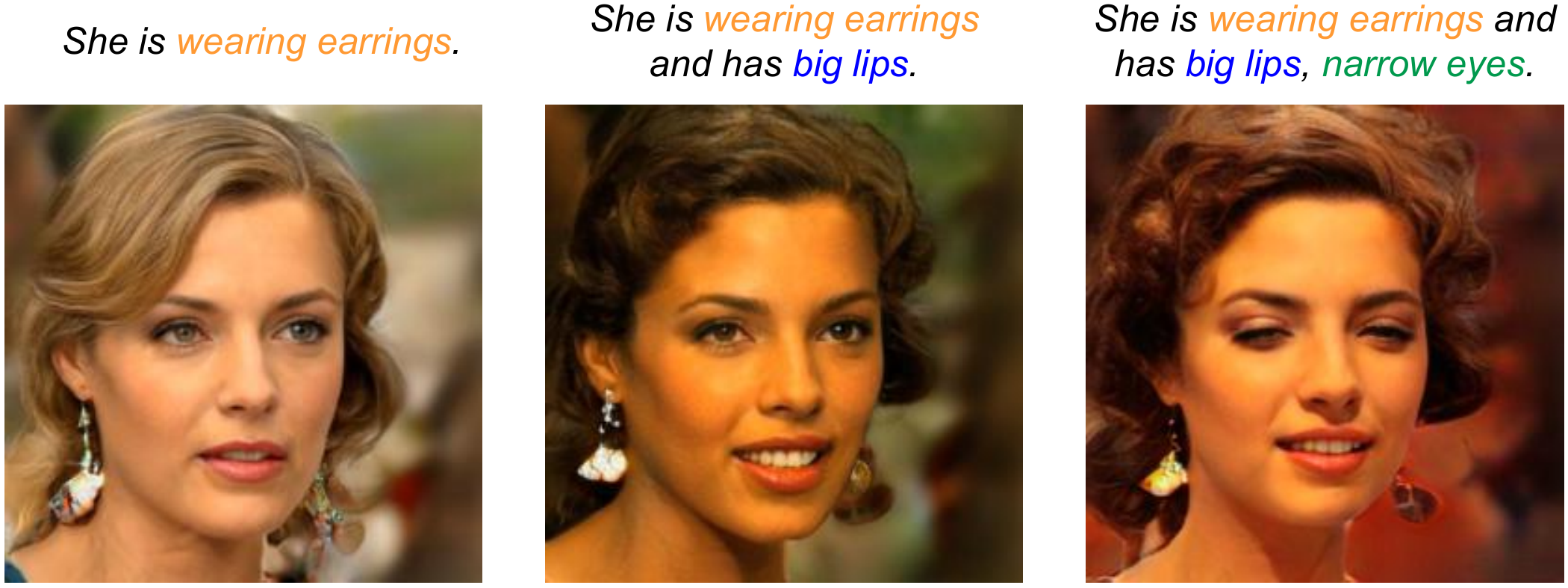}
   \caption{The performance of ACTIG when making the attribute compositions more complex.}
   \label{fig:composition_demo}
   \vspace{-3mm}
\end{figure}

The qualitative results are shown in \cref{fig:qualitative}. 
Our method, ACTIG, outperforms other state-of-the-art models in terms of image quality and text-image consistency for the CelebA-HQ and CUB datasets.
For ControlGAN \cite{li2019controllable}  and DAE-GAN\cite{ruan2021dae}{}, the output images do not have high fidelity when input prompts are complex.
TediGAN \cite{xia2021tedigan,xia2021towards} can output high-fidelity images, but sometimes the generated images and the input text do not match at all.
Lafite \cite{zhou2021lafite} has better compositional generalization, while sometimes Lafite confuses the attributes, for example, reversing the color of the bird's head and body.
We conjecture
that this is an overfitting caused by the overrepresented attribute compositions.
Compared to StyleT2I \cite{li2022stylet2i} using a pre-trained generator, the images generated by ACTIG not only accurately match the input text, but also have a higher fidelity.
\cref{fig:composition_demo} shows the introduction of new attributes to the text, ACTIG can represent all attributes and keep high image quality.
More results are shown in the supplementary material.

\begin{figure}[http]
  \centering
   \includegraphics[width=0.85\linewidth]{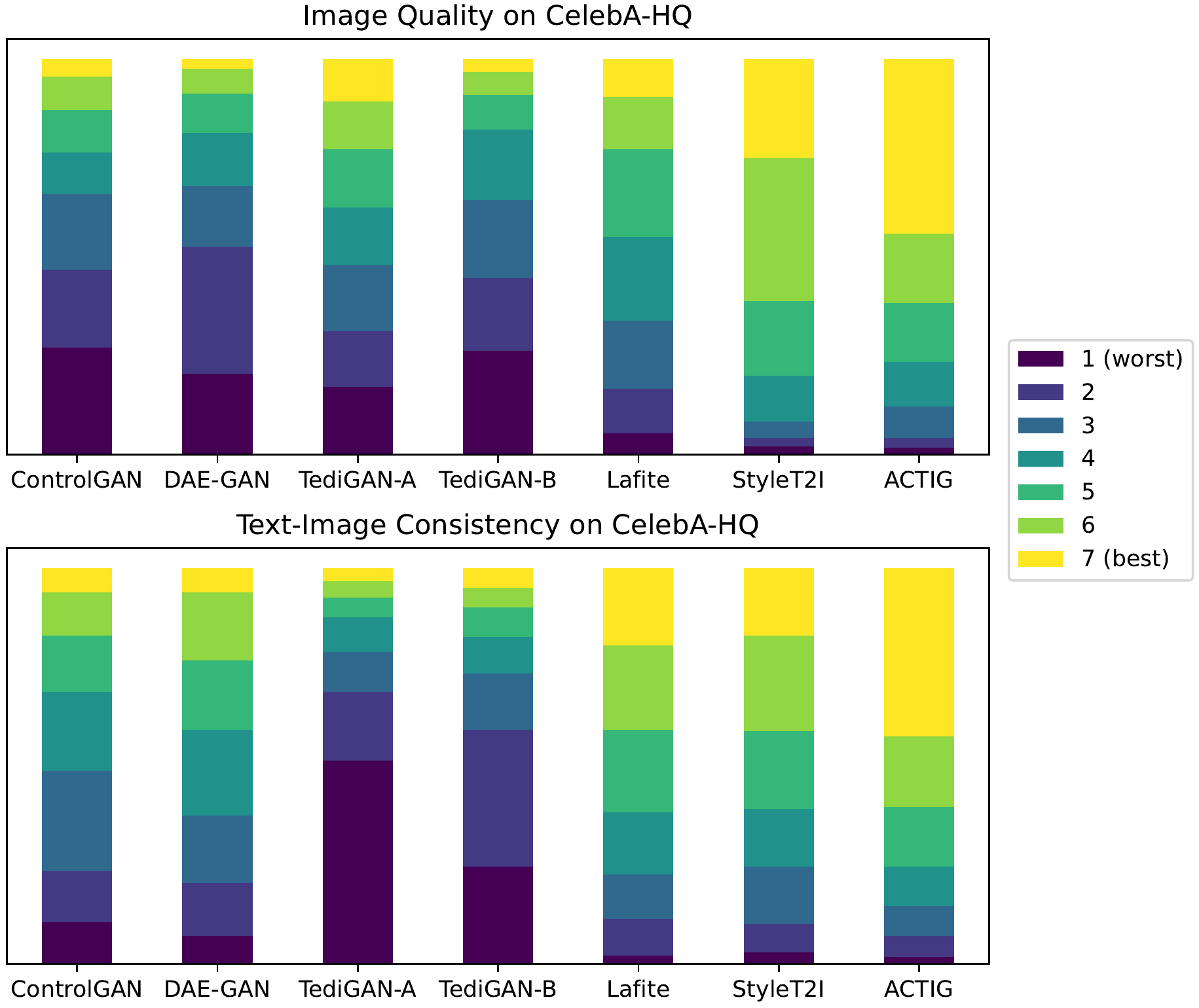}
   \caption{Score distributions of the user study on the CelebA-HQ.}
   \label{fig:user_study}
      \vspace{-2mm}
\end{figure}

\subsection{User study}
We obtain the text and images from Li \textit{et al.} \cite{li2022stylet2i} to be used in their user study and add the generated images by Lafite \cite{zhou2021lafite} and ACTIG.
$12$ participants with different backgrounds are invited to estimate 40 groups of images (20 for CelebA-HQ and 20 for CUB), each containing seven images generated by seven models with the same text.
For each group of images, participants are asked to rank and score them in terms of both image quality and text-image consistency, with a minimum score of 1 and a maximum score of 7.
The average scores of different models are shown in \cref{tab:user_study}.
Due to space limitation, only the score distributions on the CelebA-HQ are shown in \cref{fig:user_study}. 
ACTIG receives higher ranking scores for both image quality and image-text consistency.
More details and images used in the user study are included in the supplementary material.

\begin{table}[t!]
\centering
\begin{adjustbox}{width=0.49\textwidth}
\begin{tabular}{c|cc|cc}
 \hline \hline
 \multirow{2}*{Method} & \multicolumn{2}{c|}{CelebA-HQ} & \multicolumn{2}{c}{CUB} \\
  & Quality & Consistency & Quality & Consistency\\
  \hline 
ControlGAN   &  3.02 & 3.72 & 2.26& 3.50\\
DAE-GAN  &  2.89 & 4.05 & 2.50 & 2.69\\
TediGAN-A & 3.75 & 2.25 & 4.39 & 2.66\\
TediGAN-B &  2.95 & 2.75 & 3.62& 3.03\\
Lafite  &  4.22 & 4.86 & 4.41 & 4.85\\
StyleT2I   &  5.52 & 4.82 & 4.94& 5.44\\
ACTIG (Ours) & \textbf{5.65} & \textbf{5.55} & \textbf{5.95} & \textbf{5.83}\\
\hline \hline
\end{tabular}
\end{adjustbox}
\vspace{-2mm}
\caption{Average scores in term of image quality (Quality) and text-image consistency (Consistency) on the CelebA-HQ and CUB datasets. High scores represent higher performance.}
\label{tab:user_study}
\vspace{-3mm}
\end{table}

\section{Conclusion and future work}
We propose a novel attribute-centric compositional text-to-image generation framework, ACTIG, which achieves compositional generalization for both underrepresented and overrepresented attribute compositions. To improve the generalization of underrepresented attribute compositions, we introduce attribute-centric text feature augmentation and image-free training. To overcome the bias of overrepresented attribute compositions, an attribute-centric contrastive loss is proposed to learn the independent attribute distributions through adversarial training. 
ACTIG achieves
state-of-the-art results 
in terms of image fidelity and text-image consistency.
Our framework can be extended similarly to improve compositional generalization of multiple foreground entities, which is our future direction to promote the robustness of generative models.


\appendix
\section*{Appendix}
In this supplementary material, we provide additional details which help in understanding and reproducing our work.
We present the structure and training details of GAN in \cref{sec:gan_details}.
The structure details and analysis of our text-to-image mapping network are demonstrated in \cref{sec:text2img}.
We show how we implement the attribute-centric feature augmentation for two datasets in \cref{sec:feature_augmentation}, and explain how we extract the attributes when using the attribute-centric contrastive loss in \cref{sec:attribute_extraction}.
To visualize the performance of ACTIG, more qualitative results are given in \cref{sec:qualitative_results}, and we show more details of the user study in \cref{sec:user_study}.
Limitations, future work, and ethics issues are discussed in \cref{sec:discussion}.

\section{GAN details}
\label{sec:gan_details}
\subsection{GAN structure}
Our generative model is built upon StyleGAN2 \cite{karras2020analyzing} with
two modifications: (1) We adapt the original unconditional generator to a condition generator. (2) We introduce a matching discriminator for the text-image consistency.
The original discriminator is directly adopted to estimate the photo-fidelity.

\begin{figure}[http]
\centering
\includegraphics[width=1\linewidth]{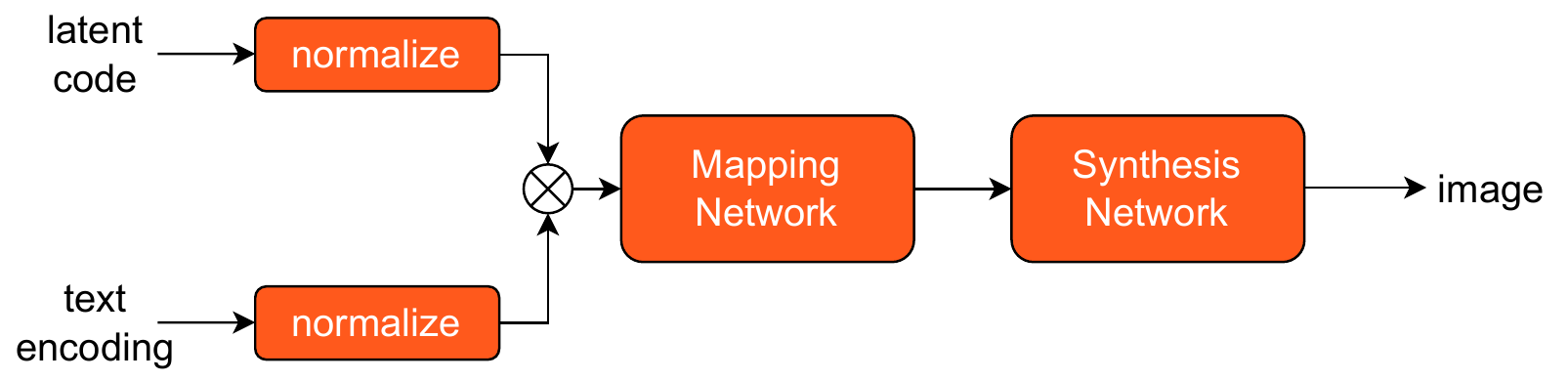}
\caption{Architecture of our generator. $ \bigotimes$ denotes the concatenation operation. }
\label{fig:supp_gan}
\vspace{-1mm}
\end{figure}

\vspace{2mm}
\noindent
\textbf{Generator.}
The original generator of StyleGAN2 is unconditional, which consists of a mapping network and a synthesis network.
To make it possible to perform the task of text-to-image generation, we normalize the text encodings from CLIP \cite{radford2021learning} and adopt the concatenation of text features and normalized latent code as the input of the mapping network.
The architecture of our generator $G$ is shown in \cref{fig:supp_gan}. 
The input dimension of the mapping network is 1024, while the output dimension is 512. 
There is no modification to the synthesis network.

\begin{figure}[ht!]
\centering
\includegraphics[width=1\linewidth]{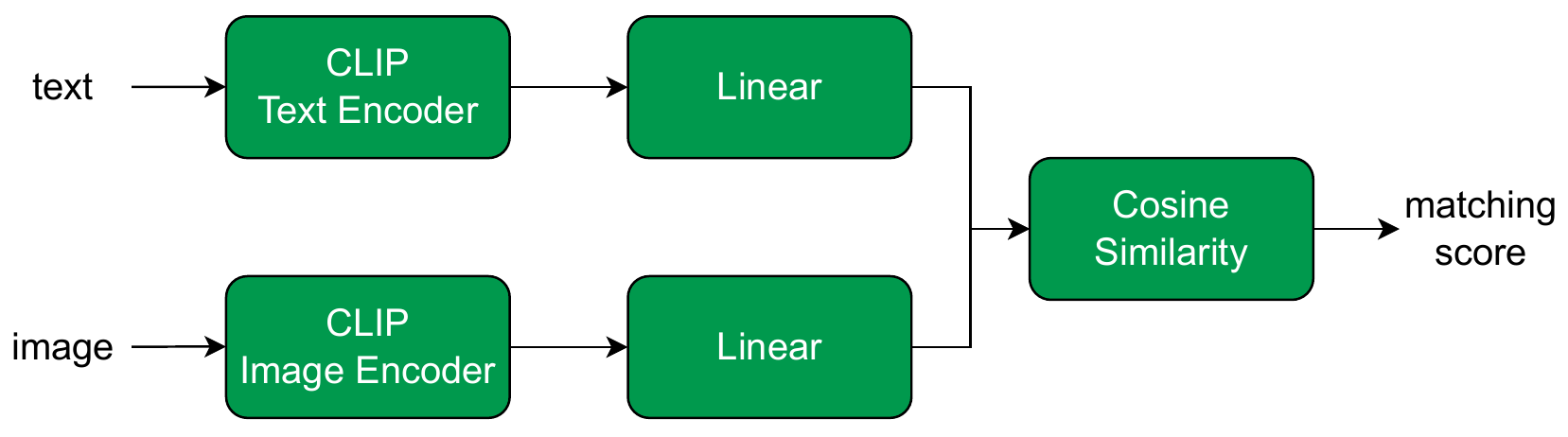}
\caption{Architecture of our matching discriminator.}
\label{fig:supp_dm}
\vspace{-1mm}
\end{figure}

\vspace{2mm}
\noindent
\textbf{Matching discriminator.}
The architecture of the matching discriminator $D_m$ is demonstrated in \cref{fig:supp_dm}.
We utilize the CLIP text encoder and image encoder to encode the input text and image, while the CLIP encoders are frozen during the adversarial training.
The output text and image encodings are forwarded to two linear layers, and the cosine similarity between them is calculated to indicate the text-image consistency.

\subsection{GAN Training}
We train the generator and discriminators from scratch on 8 GPUs, setting the batch size to 8 per GPU, while the parameters of the text-to-image mapping network, finetuned CLIP, and CLIP-A are fixed.
We use Adam \cite{kingma2014adam} optimizer with the learning
rate $2.5e^{-3}$. 
We train the generative models with $110k$ iterations for the CelebA-HQ \cite{CelebAMask-HQ} and $50k$ iterations for the CUB \cite{wah2011caltech} dataset.
We alternate between fully supervised training and image-free training, and three out of every four iterations use fully supervised training and one image-free training. 

\section{Text-to-image mapping}
\label{sec:text2img}
\begin{figure}[t!]
\centering
\includegraphics[width=1\linewidth]{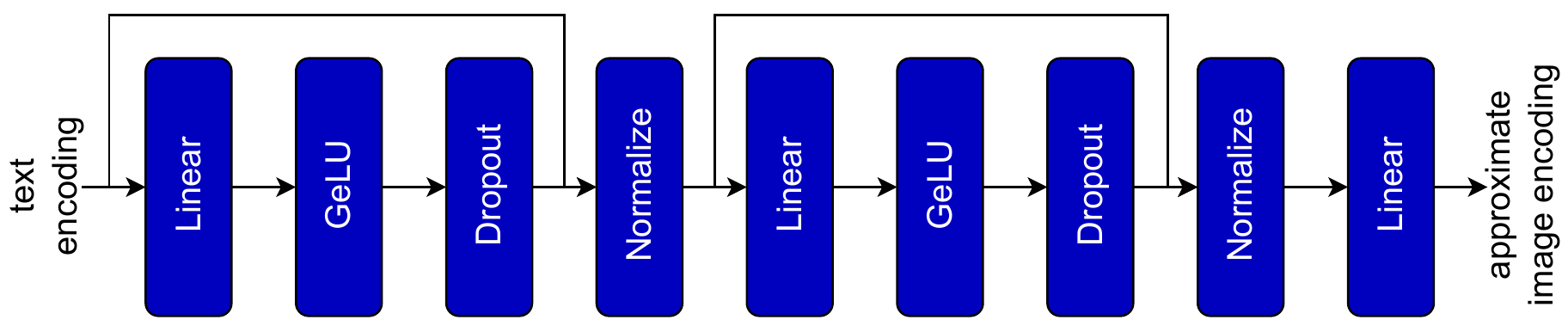}
\caption{Architecture of our text-to-image mapping network. }
\label{fig:supp_t2i}
\vspace{-1mm}
\end{figure}

A text-to-image mapping network is proposed to project CLIP text encodings into CLIP image space.
The approximate image encodings are further used in the image-free training.
The text-to-image mapping network consists of three linear transformation layers with GeLU activation and residual connections. 
The structure is shown in \cref{fig:supp_t2i}.

\begin{figure}[ht!]
\centering
\includegraphics[width=1\linewidth]{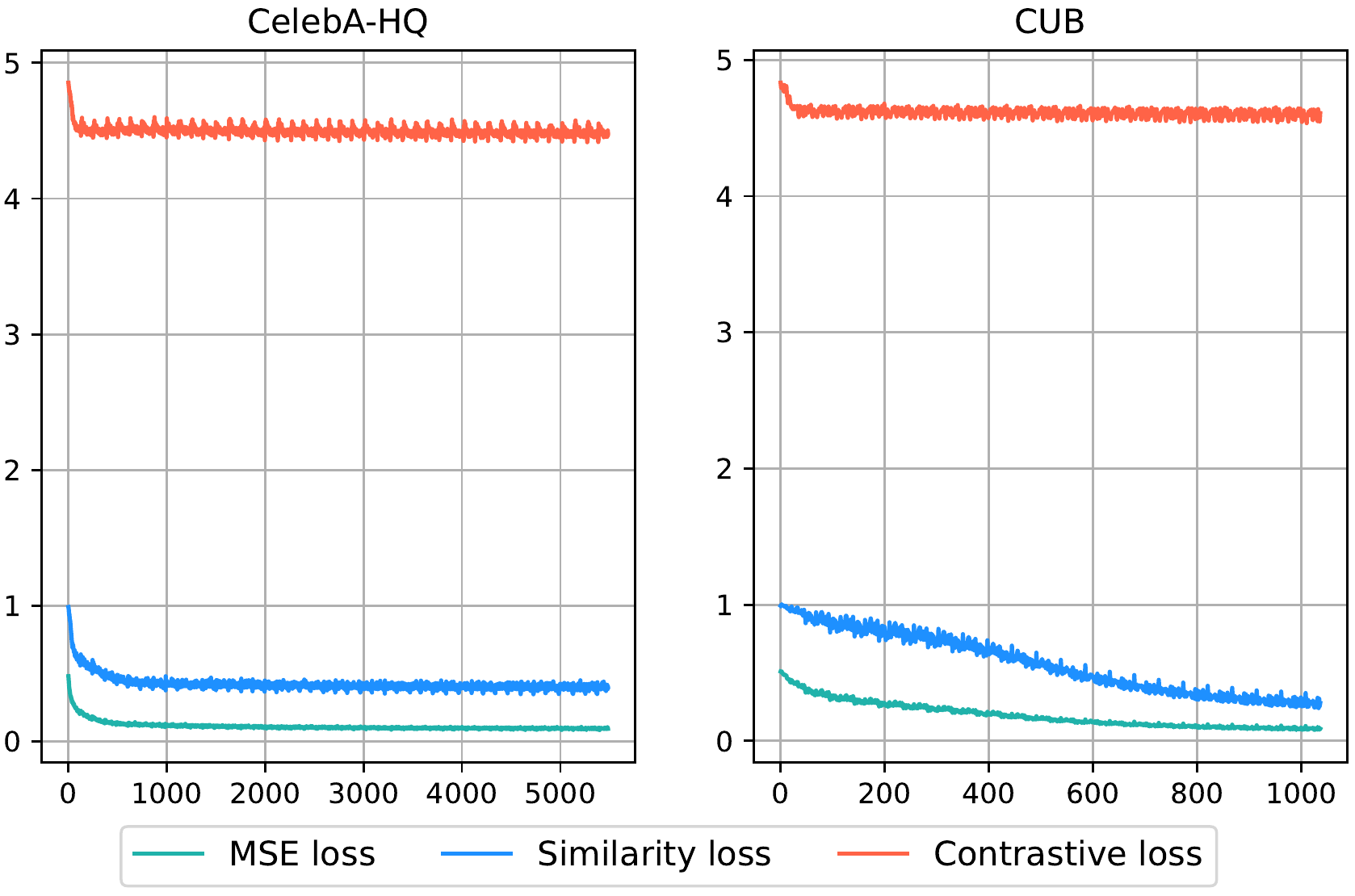}
\caption{Loss values in the training on CelebA and CUB. }
\label{fig:mapping_loss}
\end{figure}

We train the text-to-image mapping network using the text-image pairs from the training sets of CelebA-HQ and CUB for $30$ epochs with Adam. 
We use a batch size of 128 for CelebA-HQ and a batch size of 256 for CUB with the learning rate  $1e^{-4}$.
The loss values in the training are shown in \cref{fig:mapping_loss}.

To estimate the performance of the text-to-image mapping network, we compute the CLIP encodings of the text-image pairs in the test set. 
The text encodings are transformed to the approximate image encodings by our text-to-image mapping network, which are used as queries to retrieve matching CLIP image encodings. 
We adopt cosine similarity as the matching score, and the R-Precision scores are shown in \cref{tab:t2i_mapping}.
The R-Precision of CelebA-HQ is higher than that of CUB, since the visual difference between human faces is greater than the visual difference between birds.
These results are consistent with the R-precision scores of using the texts as queries.
\begin{table}[t!]
\centering
\begin{adjustbox}{width=0.25\textwidth}
\begin{tabular}{c|c}
\hline \hline
Dataset &R-Precision $\uparrow$ \\
\hline
CelebA-HQ & 78.2 \\
CUB & 52.4\\
\hline \hline
\end{tabular}
\end{adjustbox}
\caption{R-Precision of using approximate image encodings as queries to retrieve matching real image encodings.}
\label{tab:t2i_mapping}
\vspace{-2mm}
\end{table}

\section{Attribute-centric feature augmentation}
\label{sec:feature_augmentation}
We propose a novel attribute-centric feature augmentation to compensate for the feature distribution of underrepresented attribute compositions. 
The texts containing underrepresented attribute compositions are generated first, while the requirement for images is weakened by mapping CLIP text features to CLIP image space.
The augmented texts are encoded by  CLIP text encoder, while the approximate image encodings are computed by our text-to-image mapping network.
For the CelebA-HQ and CUB datasets, text augmentation is performed in two different ways.
\subsection{CelebA-HQ dataset}
The original captions in the CelebA-HQ dataset are generated by probabilistic context-free grammar based on the known attribute labels.
We first collect the keywords of these attribute labels and group them into two categories:
\begin{itemize}
\item Gender: \textit{he, man, she, woman.}
\item Appearance: \textit{arched eyebrows, bags under eyes, bangs, big lips, big nose, black hair, blond hair, brown hair, bushy eyebrows, double chin, goatee, gray hair, high cheekbones, mouth slightly open, mustache, narrow eyes, oval face, pale skin, pointy nose, receding hairline, rosy cheeks, sideburns, straight hair, wavy hair, bald, chubby, smiling, young, eyeglasses, heavy makeup, earrings, hat, lipstick.}
\end{itemize}
We synthesize $10k$ augmented prompts which are further used in the image-free training. 
When generating a prompt, the gender is first randomly determined, and then we randomly sample two to six attributes of appearance.
The sampled attributes are composed into a prompt according to the syntax rules,
for example, \textit{``the man has gray hair and straight hair, and he wears lipstick''}.

\begin{figure*}[ht!]
\centering
\includegraphics[width=0.95\linewidth]{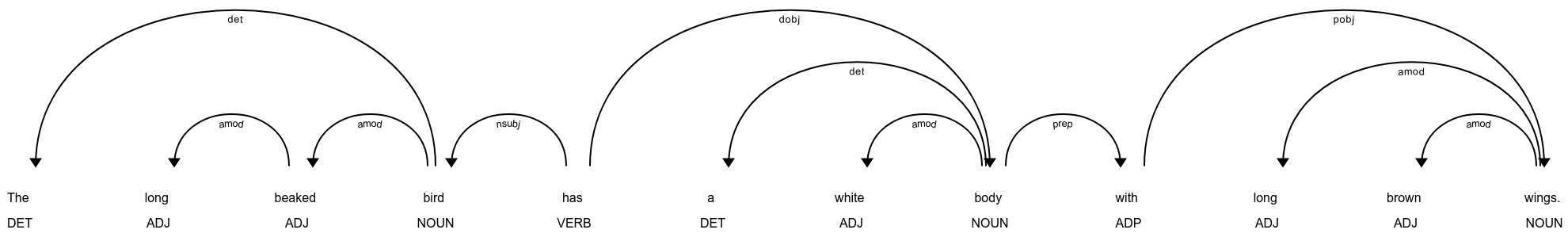}
\caption{Visualization of the dependency parse for the text, \textit{``the long beaked bird has a white body with long brown wings''}.
The dependency matcher can localize the attributes in the text with the dependency matcher patterns.}
   \label{fig:spacy}
   \vspace{-1mm}
\end{figure*}

\subsection{CUB dataset}
\label{sec:augment_cub}
Different from CelebA-HQ, the captions in the CUB dataset are written artificially.
In order to generate the captions with underrepresented attribute compositions, we first design a attribute parser based on the dependency matcher implemented in spaCy  (see \cref{fig:spacy}).
We use the attribute parser to extract attributes from the prompts in the training set, while keeping the noun in the attribute unchanged and replace the adjective.
We divide the adjectives appearing in the dataset into color and shape according to \cite{park2021benchmark}, and create the attribute library:
\begin{itemize}
\item Color: \textit{brown, white, yellow, dark, gray, grey, black, red,  rusty, beige, maroon, orange, green, iridescent, lime, pink, pale, purple, blue, taupe, gold, bronze, amber, magenta, silver, lightbrown, flittery, violet, teal, crimson, olive, creamy,  metallic, azure, turquoise, indigo, chocolate, ruby, bluegreen, mauve, tawny, ivory, ash, khaki, scarlet, cyan, lemon, rosy, coppery, peachy, blond, earthtone, inky, opalescent, tan.}
\item Shape: \textit{small, little, short, pointy, narrow, large, long, straight, medium, curved, pointed, thin, tiny, sharp, curving, skinny, stout, chubby, tall.}
\end{itemize}
We also synthesize $10k$ augmented prompts for the CUB dataset. 
During the synthesis process, the colors in the text are replaced with other randomly sampled colors and the shapes are also replaced with the new shapes.
For example, based on the prompt in the training set, \textit{``the long beaked bird has a white body with long brown wings''}, we replace the attributes and obtain the new prompt, \textit{``the tiny beaked bird has a blond body with tiny blue wings''}.

\section{Attribute extraction}
\label{sec:attribute_extraction}
To perform the attribute-centric contrastive loss, an attribute is randomly sampled from the text in each iteration of training.
For the CelebA-HQ dataset, we extract the attributes in the sentence using string matching, since the sentences are generated based on the known attribute labels. 
For the CUB dataset, the attributes are localized by the attribute parser introduced in \cref{sec:augment_cub}.

\section{Additional qualitative results}
\label{sec:qualitative_results}
\begin{figure*}[ht!]
\centering
\includegraphics[width=1\linewidth]{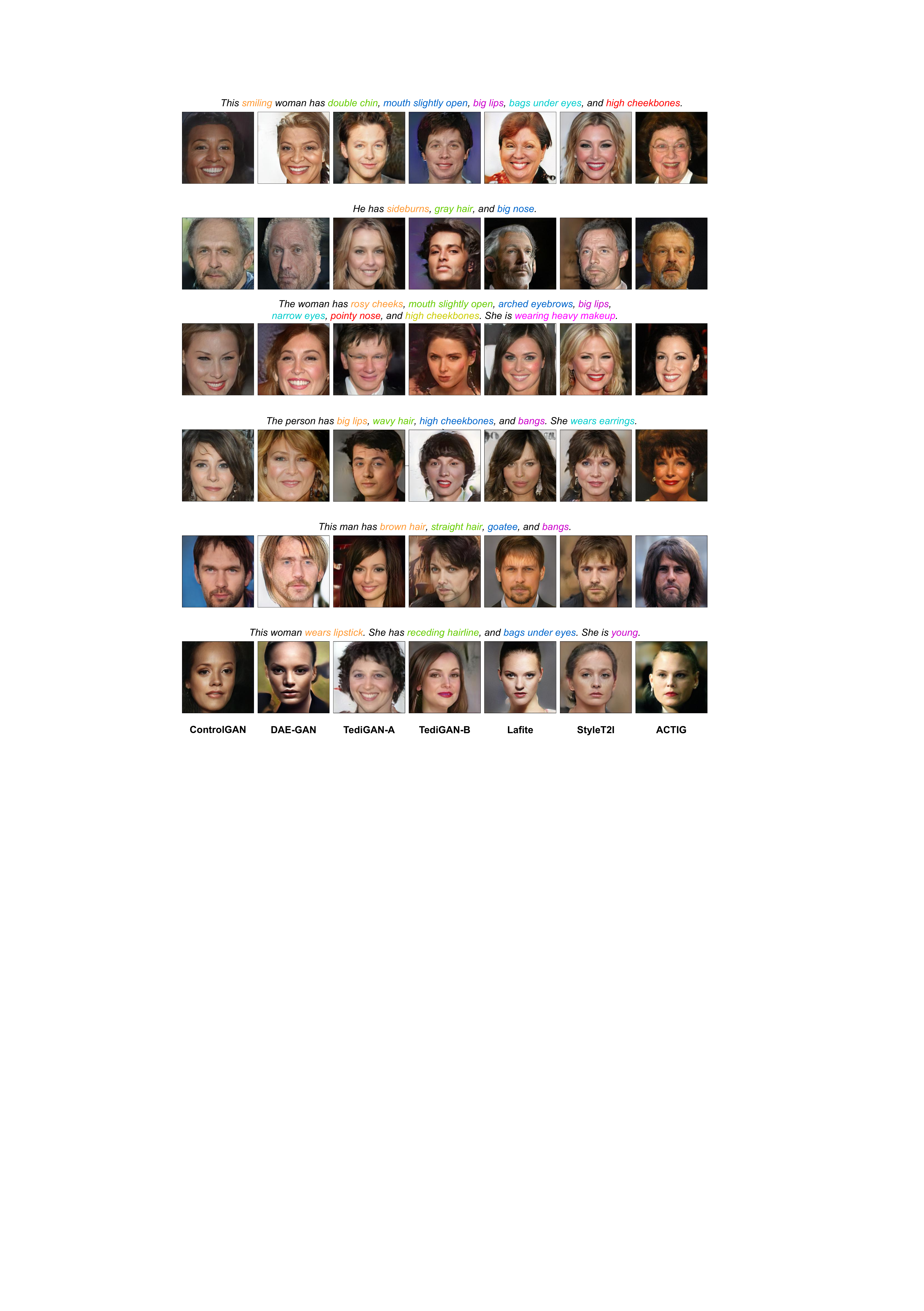}
\caption{ Additional qualitative results on the CelebA-HQ dataset.}
   \label{fig:supp_celeba}
   \vspace{-3mm}
\end{figure*}

\begin{figure*}[ht!]
\centering
\includegraphics[width=1\linewidth]{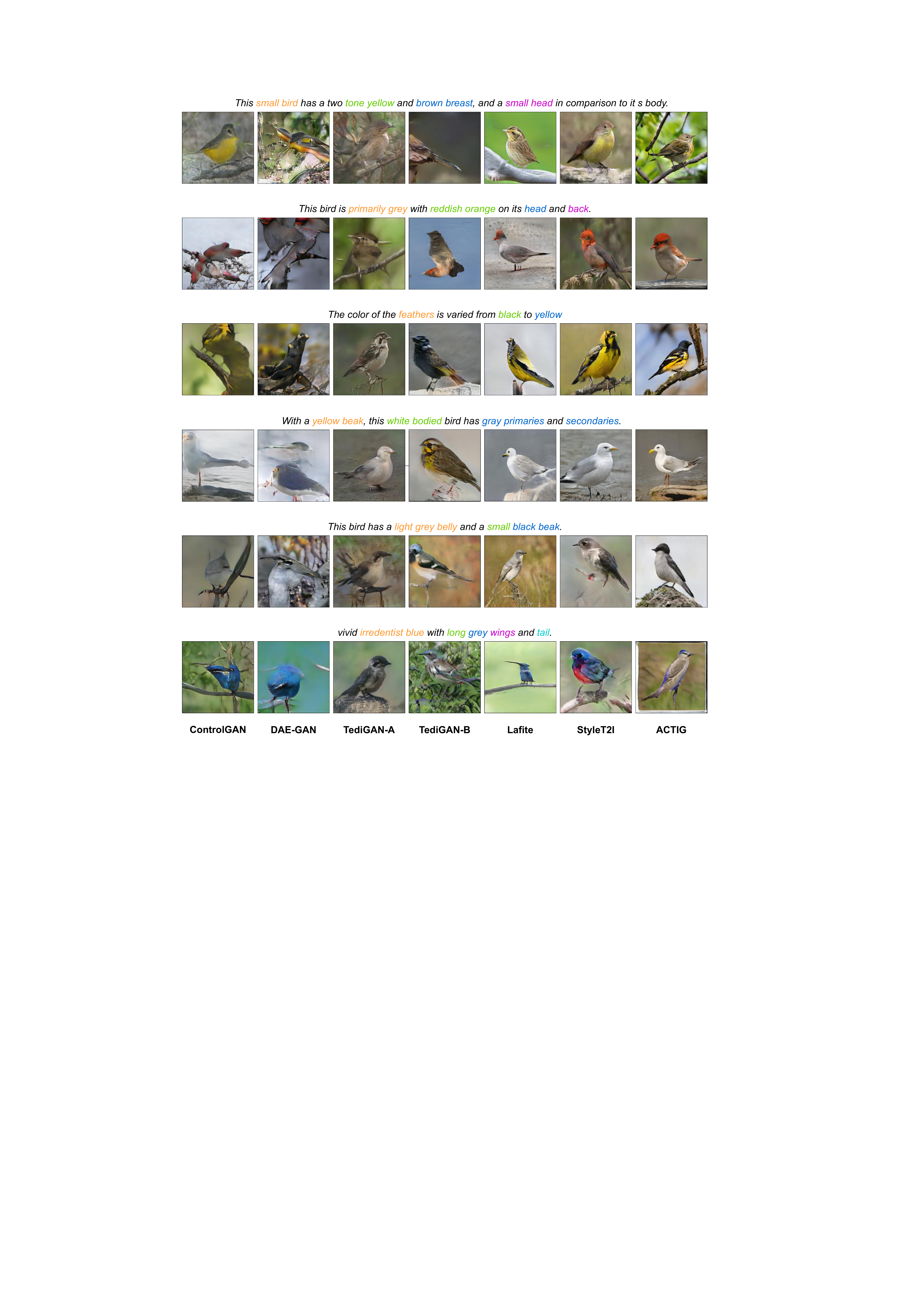}
\caption{Additional qualitative results on the CUB dataset.}
   \label{fig:supp_cub}
   \vspace{-3mm}
\end{figure*}

The additional qualitative results for the CelebA-HQ and CUB datasets are respectively shown in \cref{fig:supp_celeba} and \cref{fig:supp_cub}. 
The attribute compositions in the input prompts have not been seen in the training sets.
To better visualize the attribute compositions, we use different colors to highlight the attributes in the prompts.
Note that the gender attributes are not colored for the CelebA-HQ dataset, \textit{e.g.}, \textit{``she''} and \textit{``this man''}.
Overall, our model has outstanding performance in term of image quality and text-image consistency in both Celeb-HQ and CUB.
Compared to another compositional text-to-image generation model, StyleT2I \cite{li2022stylet2i}, our model can generate more realistic images.
We conjecture that the reason is that our latent space is optimized, while StyleT2I is performed on a pre-trained generator.


\section{User study}
\label{sec:user_study}
\begin{figure*}[ht!]
\centering
\includegraphics[width=1\linewidth]{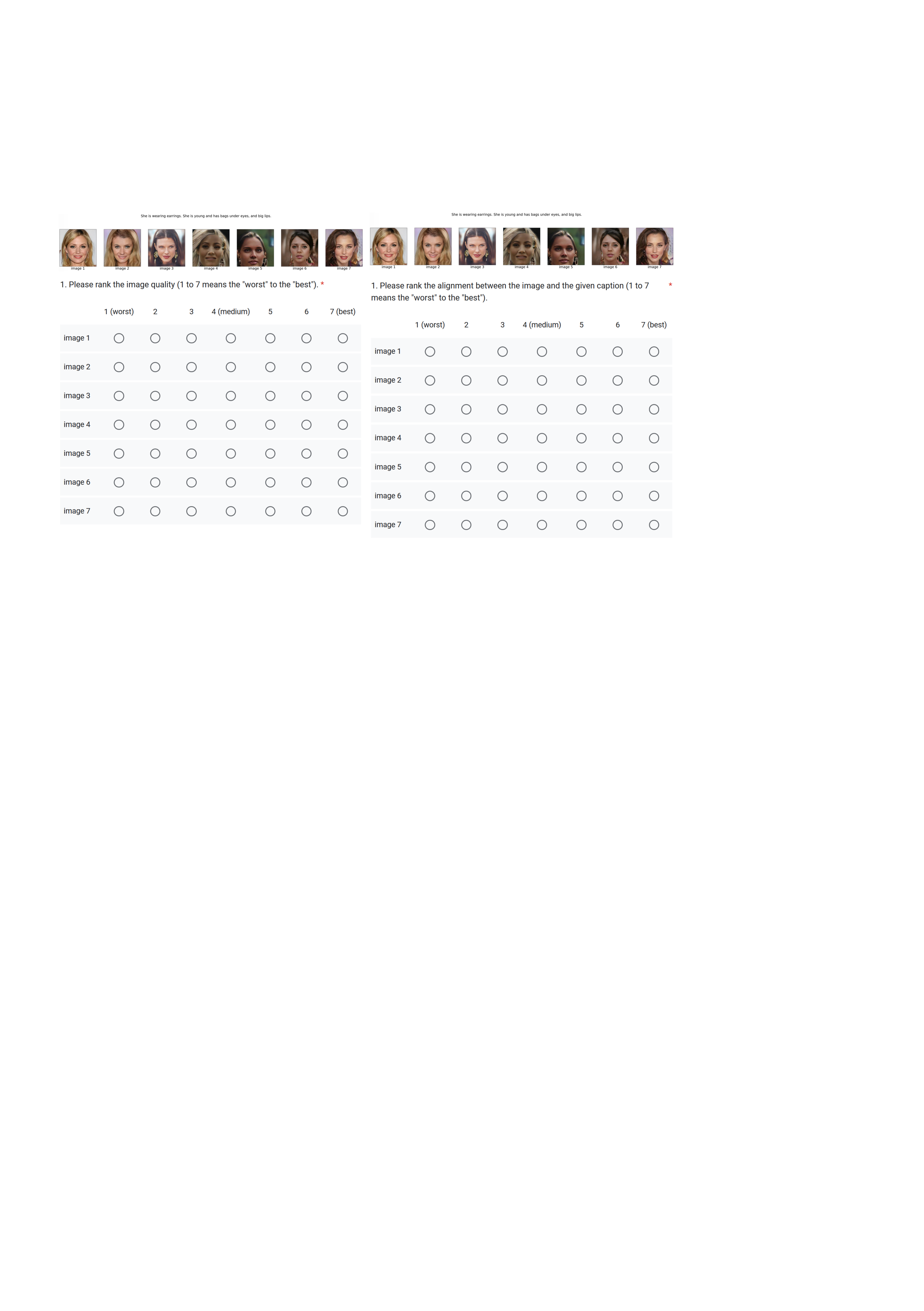}
\caption{User study interface for image quality and text-image consistency evaluation: one text-image group for CelebA-HQ.}
   \label{fig:user_study_demo_celeba}
\end{figure*}

We adopt the images used for the user study in StyleT2I and add the images generated by Lafite \cite{zhou2021lafite} and ACTIG.
There are 40 text-image groups in the user study, 20 for CelebA-HQ and 20 for CUB.
For each group, 7 images generated by 7 different models and the corresponding text are shown.
We request 12 participants to rank the given images in term of image quality and text-image consistency.
One text-image group for CelebA-HQ is shown in \cref{fig:user_study_demo_celeba}.
Each participant sees two types of questions:
\begin{enumerate}
    \item Rank the alignment between the image and the given caption. 
When answering this type of question, the participant is asked to focus on the semantic similarity between the caption and image.
    \item Rank the image quality (how close the generated image is to the real image). When answering this type of question, the participant is asked to focus on the quality of the image (\eg, fidelity, blur, artifacts) instead of the semantic similarity with the caption. 
\end{enumerate}
In both types of questions, 1 means the "worst", and 7 represents the "best".
Note that one score can only be assigned to one image. 

The average score distributions of the generative models on CelebA-HQ and CUB are shown in \cref{fig:user_study}.
For both datasets, ACTIG receives higher ranking scores in term of image quality and image-text consistency. 


\begin{figure*}[ht!]
\centering
\includegraphics[width=1\linewidth]{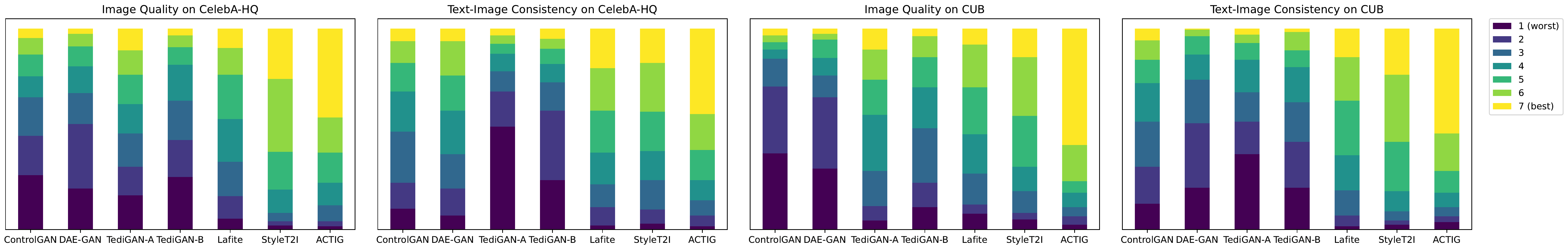}
\caption{Score distributions of the user study on CelebA-HQ and CUB.}
   \label{fig:user_study}
   \vspace{-1mm}
\end{figure*}

\section{Discussion}
\label{sec:discussion}

\subsection{Limitations and future work}
Although ACTIG achieves state-of-the-art results in terms of image fidelity and text-image consistency, there are still some limitations.
We introduce an attribute-centric feature augmentation and an image-free training to compensate for the data distribution. 
However, the image features provided by the text-to-image mapping network can not represent exact visual appearance.
It could cause that the generated images containing underrepresented attribute compositions match the input text, but the image quality is not high enough.
A potential approach is to use external images to update the fidelity discriminator in the image-free training.
In addition, our attribute extraction (for CUB) is based on the dependency matcher.
For some very complex sentences the attribute parser cannot extract all attributes accurately.
ACTIG focuses on the attribute centric compositional text-to-image generation.
A similar extension to our approach can improve the compositional generalization of multiple foreground entities.

\subsection{Ethics statement}
\label{sec:ethics}
As the use of machine learning  in everyday life grows, it is relevant to consider the potential social impact of our work.
Our work has the potential to be used for deep fake.
Since our model can generate high-fidelity images with specific attributes, this even makes deep fake more flexible.
On the other hand, our attribute-centric generative model is less affected by overrepresented attribute compositions in the dataset and can generate the images that match the given text.
Therefore, our work contributes to the elimination of bias in generative models.
{\small
\bibliographystyle{ieee_fullname}
\bibliography{egbib}
}

\end{document}